\documentclass[runningheads,a4paper]{llncs}
\usepackage{graphicx}
\usepackage{subfigure}
\usepackage{amsmath, amssymb}
\usepackage{tabularx}
\usepackage{booktabs}
\usepackage{color}
\usepackage{xcolor}
\usepackage{paralist}
\usepackage{cite}
\usepackage{multirow}
\usepackage{tabularx}
\usepackage{CJKutf8}
\usepackage{float}

\begin{document}
\mainmatter  

\title{BERT Meets Chinese Word Segmentation}

\titlerunning{Bert for CWS}

%
%
\author{Haiqin Yang
}
\authorrunning{} 
\toctitle{Lecture Notes in Computer Science}
%
\institute{Meitu \and 
Department of Computing, The Hang Seng University of Hong Kong
\email{haiqin.yang@gmail.com}
}

%


%
%

\maketitle

\begin{abstract}
Chinese word segmentation (CWS) is a fundamental task for Chinese language understanding.  Recently, neural network-based models have attained superior performance in solving the in-domain CWS task.  Last year, Bidirectional Encoder Representation from Transformers (BERT), a new language representation model, has been proposed as a backbone model for many natural language tasks and redefined the corresponding performance.  The excellent performance of BERT motivates us to apply it to solve the CWS task.  By conducting intensive experiments in the benchmark datasets from the second International Chinese Word Segmentation Bake-off, we obtain several keen observations.  BERT can slightly improve the performance even when the datasets contain the issue of labeling inconsistency.  When applying sufficiently learned features, Softmax, a simpler classifier, can attain the same performance as that of a more complicated classifier, e.g., Conditional Random Field (CRF).  The performance of BERT usually increases as the model size increases.  The features extracted by BERT can be also applied as good candidates for other neural network models. 
\end{abstract} 

\section{Introduction}
Chinese word segmentation (CWS), i.e., dividing text into words, is a key preprocessing step for Chinese language understanding~\cite{DBLP:journals/corr/abs-1901-06079}.  This task can be modeled as a token tagging task or a character-based sequence labelling task~\cite{DBLP:conf/emnlp/MaGW18}. 

Recently, neural network models have been applied to solve this task with less effort in feature engineering~\cite{DBLP:conf/acl/CaiZ16,DBLP:conf/acl/PeiGC14,DBLP:conf/emnlp/ChenQZLH15,DBLP:conf/emnlp/MaGW18}.  For example, in~\cite{DBLP:conf/acl/PeiGC14}, Max-Margin Tensor Neural Network (MMTNN) has been proposed to model interactions between tags and context characters.  In~\cite{DBLP:conf/acl/ChenQZH15}, gated recursive neural network (GRNN) is exploited to model the combination of characters for CWS.  In~\cite{DBLP:conf/emnlp/ChenQZLH15}, four different architectures of long short-term memory (LSTM) are presented and evaluated to test the performance of CWS.   In~\cite{DBLP:conf/ijcnlp/WangX17}, convolutional neural network is incorporated with word embeddings for CWS.  A thorough investigation of LSTM for CWS is presented in~\cite{DBLP:conf/emnlp/MaGW18}.  The essential of these methods boils down to two issues: 1) how to represent each character in an effective way? 2) how to absorb transition between characters to utilize contextual information? 

\begin{CJK}{UTF8}{gbsn}
\begin{figure}
\begin{tabular}{@{~~~}l@{~~~~~~~~~}cccccccccccll}
\hline\hline
\textit{Source tags:} & \multicolumn{1}{c|}{S} & \multicolumn{1}{c|}{S} & \multicolumn{2}{c|}{B\quad~E} & \multicolumn{2}{c|}{B~~E}  & \multicolumn{1}{@{~~~}c@{~~~}}{S} &
\\ \hline
\textit{Source:} & \multicolumn{1}{l|}{Confluence~~} & \multicolumn{1}{c|}{~于~~} & \multicolumn{2}{c|}{~~2004~年~~} & \multicolumn{2}{c|}{~首~发~~} & \multicolumn{1}{@{~~~}c@{~~~}}{。}& 
\\\hline
\textit{Meaning:} &  \multicolumn{10}{l}{Confluence was first released in 2004.}
\\\hline\hline
\textit{BERT tags:} & \multicolumn{5}{c|}{B\qquad~M\qquad~M\qquad~E\qquad\qquad} & \multicolumn{1}{@{~}c@{~}|}{S} & \multicolumn{2}{c|}{B\quad~E} & \multicolumn{2}{c|}{B~~E}  & \multicolumn{1}{@{~~~}c@{~~~}}{S} &\\\hline
\textit{BERT: } & \multicolumn{5}{l|}{con\quad\#\#f\,l\quad\#\#ue\quad\#\#nce~~}  & \multicolumn{1}{c|}{~于~~} & \multicolumn{2}{c|}{~2004~年~~} & \multicolumn{2}{c|}{~~首~发~~} & \multicolumn{1}{@{~~~}c@{~~~}}{。} & ~~~~
\\\hline\hline
\end{tabular}
\caption{An example of Chinese words tagging: a slight difference lies in the source tags and the BERT tags for handling the English words, see detailed description in the text.
\label{fig:example}}
\end{figure}
\end{CJK}

To address the above problems, Yang et al. have learned pretrained character/word embeddings for characters, character bigrams, and words from rich external resources and shown significant error reduction in CWS~\cite{DBLP:conf/acl/YangZD17}.  GRNN, LSTM, and CNN have been applied to model the coherence in segmented sentences~\cite{DBLP:conf/acl/ChenQZH15,DBLP:conf/emnlp/ChenQZLH15, DBLP:conf/ijcnlp/WangX17}, but they require to specify a fixed context window, which lacks the flexibility of capturing the contextual information sufficiently.  In~\cite{DBLP:conf/acl/CaiZ16}, the limitation of fixed size context windown is overcome by employing a gated combination neural network over characters for word representation generation with an LSTM scoring model for segmentation.  The word segmenter is further sped up via greedy search~\cite{DBLP:conf/acl/CaiZZXWH17}.  However, these methods do not exploit sufficient out-domain resources and may restrict the potential power to improve the performance.

Nowadays, huge language models from unsupervised learning of abundant out-domain resources, such as ELMo~\cite{DBLP:conf/naacl/PetersNIGCLZ18} and OpenAI GPT~\cite{OpenAI_GPT}, have demonstrated the promising of utilizing information learned from out-domain resources.   Especially, Bidirectional Encoder Representation from Transformers (BERT)~\cite{DBLP:conf/naacl/DevlinCLT19} has been proposed and redefined the state of the art for eleven natural language processing tasks.  The outstanding performance of BERT and its capability to capture the contextual in the text motivates us to apply it as preprocessing step to extract features for CWS.  

In this paper, we try to understand what performance BERT can be attained in solving the CWS task in various aspects: 
\begin{compactitem}
\item Can BERT continue improving the performance of the CWS task? 
\item What is the trade-off between character representations and classifiers? 
\item What is the effect of model size of BERT for the CWS task? 
\item What is the effect of BERT features working as an ELMo-like representation for the CWS task? 
\end{compactitem}
By conducting extensive experiments on two benchmark CWS datasets, we get several first-hand and key observations about BERT and demonstrate its advantages in solving the CWS task in Sec.~\ref{sec:exp}. 

\begin{figure*}[!hbt]
\centering
\subfigure[Transformer]{\includegraphics[width=0.2\textwidth]{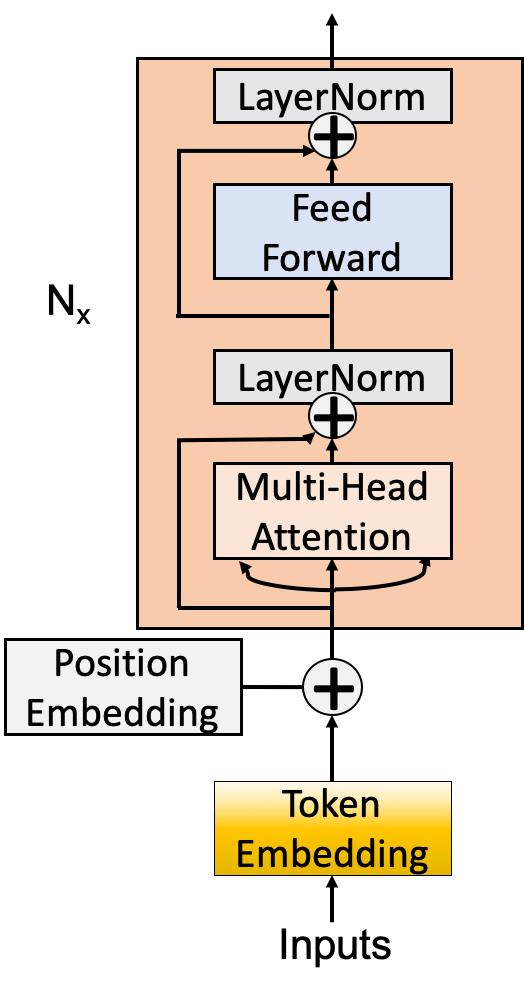}\label{fig:transformer}}
\subfigure[Chinese Word Segmentation]{\includegraphics[width=0.78\textwidth]{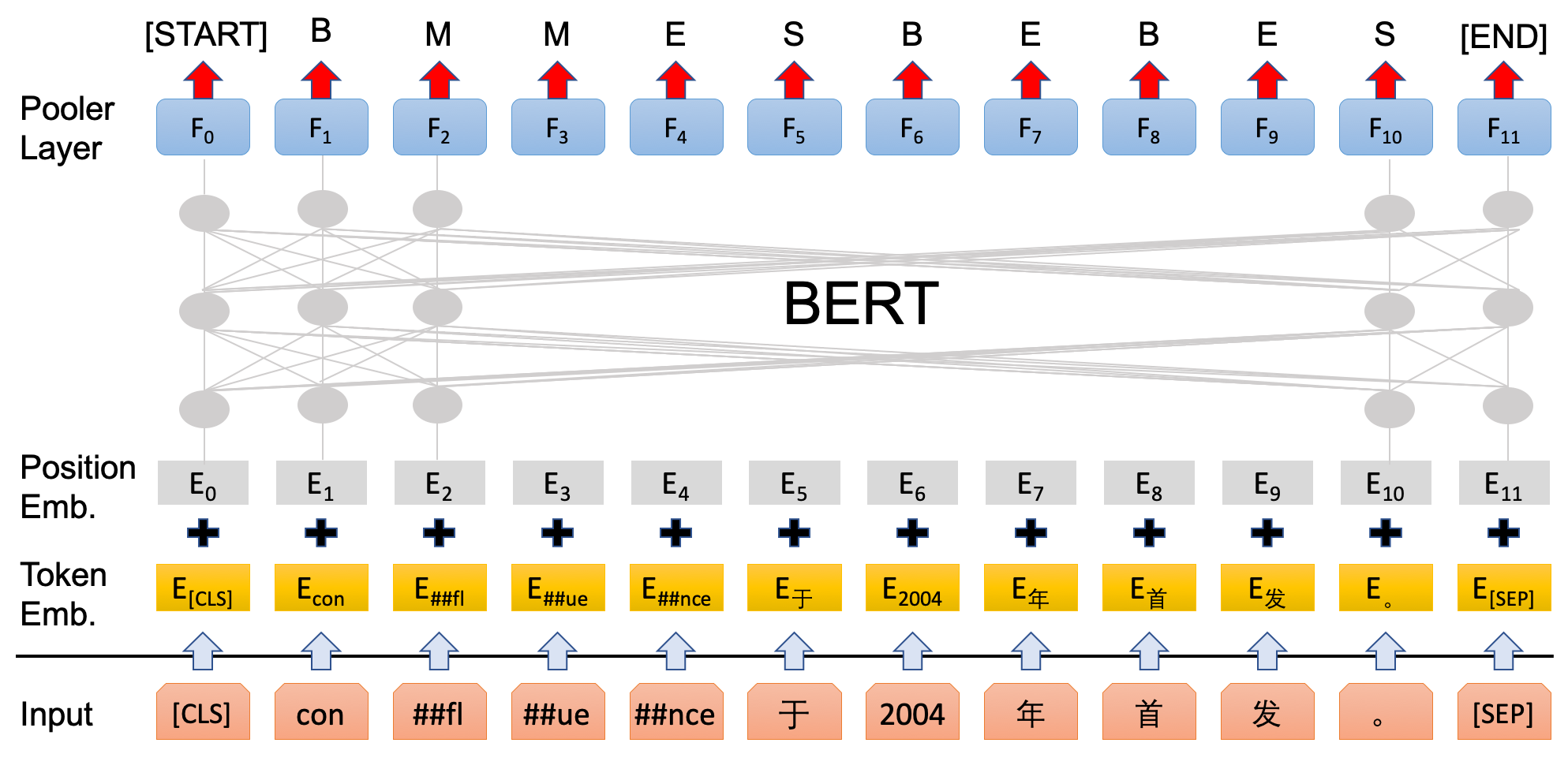}\label{fig:CWS_procedure}}
\caption{Architecture of Transformer and BERT for the CWS task.}
\end{figure*}

\section{Background and Architecture}
\label{sec:bg}
In the following, we first define the problem of CWS task and the basic concept of tokenization.  After that, we present the BERT architecture and apply it for the CWS task. 

\subsection{Problem Definition}
The problem of Chinese word segmentation is defined as follows: given an input sentence with $m$ characters $s = c_1 c_2 \ldots c_m$, where $c_i$ denotes the $i$-th character, the segmentor is to assign each character $c_i$ with a label $l_i$, where $l_i \in \{B, M, E, S\}$~\cite{DBLP:journals/ijclclp/Xu03}.  The label $B$, $M$, $E$ and $S$ represent the begin, middle, end of a word and single character word, respectively.


In this paper, we adopt the WordPiece tokenization~\cite{DBLP:journals/corr/WuSCLNMKCGMKSJL16}, which is adopted in the implementation of BERT.  The WordPiece tokenization makes no difference in handling Chinese characters, only with a slightly difference when handling English words or digits.  For example, as shown in Fig.~\ref{fig:example}, the English word, Confluence, is separated into four parts, con, \#\#fl, \#\#ue, and \#\#nce, which yield the corresponding BERT tag of BMME, rather than the source tag of S.  For the word, \begin{CJK}{UTF8}{gbsn}``2004年" (in 2004), the corresponding BERT tag is BE, where ``2004" is annotated by B and ``年" (year) is annotated by E.  It is fortunate that 2004 is deemed as a whole word, rather than the case of ``２４２４位通信院士'' (2424 communication fellow), segmented as ``２４２４ 位 通信 院士", which is tokenized as ２４, \#\#２, \#\#４, 位, 通, 信, 院, 士, respectively.  Hence, , the source tags of ``２４２４位通信院士'' are denoted by SSBEBE and its BERT tags will be changed to BMESBEBE, respectively.  In the process, if a character does not appear in the vocabulary, it is marked as the special token, [UNK].  If there are two consecutive English words, we add a special token, [unused1], to replace the space in the word.\end{CJK}  This makes our procedure more difficult than previous methods~\cite{DBLP:conf/acl/CaiZ16,DBLP:conf/acl/CaiZZXWH17,DBLP:conf/ijcnlp/WangX17}, which treat continuous digits and English characters as a single token.

As shown in Fig.~\ref{fig:CWS_procedure}, two special tokens, [CLS] and [SEP], are additionally added to denote the beginning and the end of each sentence, respectively, and yield the corresponding output tokens, [START] and [END].  These two output tokens are necessary tokens for Conditional Random Field (CRF), which needs to model the dependance between labels.

\subsection{BERT}
The essential architecture of BERT is a multi-layer bidirectional Transformer encoder to learn representations by conditioning both the left and right contexts in all layers (see Fig.~\ref{fig:transformer} for an illustration, or details in~\cite{DBLP:conf/nips/VaswaniSPUJGKP17}).  The original pretrained representation is trained via masked language models on BooksCorpus (800M words)~\cite{DBLP:conf/iccv/ZhuKZSUTF15} and English Wikipedia (2,500M words) while the multilingual model is trained on the XNLI dataset, a total of 112,500 annotated pairs in 15 languages~\cite{DBLP:conf/emnlp/ConneauRLWBSS18}.   


In terms of sequence labeling tasks, given a sequence of $m$ characters, $s = c_1 c_2 \ldots c_m$, we can formulate BERT's architecture as follows: 
\begin{eqnarray}
h_i^0 &=& W_ec_i+W_p, \\
h_i^l &=& {\rm transformer\_block}(h_i^{l-1}),\quad l=1,\ldots, L, \\\label{eq:output}
y_i^{\rm (BERT)} &=& {\rm classifier}(W_oh_i^L+b_o),
\end{eqnarray}
where $c_i$ is the $i$-th token, $W_e$ is the weight for the embedding layer, $W_p$ is the positional encoding.  Here, we additionally add the special token, [START], as $c_0$ and [END] as $c_{m+1}$.  $L$ is the number ${\rm transformer\_block}$ layers, which consists of self-attention and fully connected layers~\cite{DBLP:conf/nips/VaswaniSPUJGKP17}.  $W_o$ and $b_o$ is the weight matrix and the bias for the output layer, respectively.  The ${\rm classifier}$ can be CRF or Softmax. 

In our adopted ${\rm BERT}_{\rm BASE}$, $L=12$.  $W_e\in \mathbb{R}^{H\times |D|}$, where $H=768$ and $|D|=21,128$ by applying the ${\rm BERT}_{\rm BASE}$ model on the Chinese set, which consists of 21,128 tokens in both English, Chinese, emoji and some special symbols.  The positional coding $W_p\in \mathbb{R}^{H\times 1}$ with the maximum sequence length being 512.  The output weight matrix $W_o\in \mathbb{R}^{T\times H}$ and $b_o\in \mathbb{R}^{T\times 1}$, where $T$ is the number of output tags, i.e., $6$ in our test.



\begin{table}
\caption{Statistics of datasets.  '\#' symbol stands for the term 'the number of'. \label{tb:dataset} }
\begin{tabular}{@{~}l@{~}|l@{~~~}c@{~~~}c@{~~~}c@{~~~}c@{~~~}c@{~~~}c@{~~~}c@{~~~}}
\toprule
\multirow{2}{*}{Dataset} & \multirow{2}{*}{~~Part} & \multirow{2}{*}{\#Sent.} & \multirow{2}{*}{\#Words} & \#Chi. & \#Eng. & \multirow{2}{*}{\#Digits} & \multirow{2}{*}{\#Chars} & \multirow{2}{*}{~~OOV} \\ 
& & & & Words & Words & &\\
\midrule
\multirow{2}{*}{MSR} &~~Train & 87K & 2,368K & 2,350K & 1,154 & 18K &  4,050K & 
{~~1,991/}\\\cline{2-8}
& ~~Test & 4K  & 107K & 106K & 66 & 697 & 184K & ~~0.023\\\hline
\multirow{2}{*}{PKU} & ~~Train & 19K & 1,110K & 1090K & 443 & 20K & 1,826K & 
{~~2,863/} \\\cline{2-8}
& ~~Test & 2K & 104K & 102K & 28 & 2K & 173K & ~~0.052
\\\bottomrule
\end{tabular}
\end{table}

\section{Experiments}\label{sec:exp}
\begin{table}\caption{Comparison with previous models\label{tb:main_rs}}
\begin{tabular}{@{~}l@{~}|@{~}c@{~~~}c@{~~~}c@{~~}c@{~~~}c@{~~}c@{~~}|@{~~}c@{~~~}c@{~~~}c@{~~}c@{~~}@{~~~}c@{~~~}}
\toprule
\multirow{3}{*}{Method}  & \multicolumn{5}{c}{MSR} && \multicolumn{5}{c}{PKU} \\\cline{2-12}
& \multicolumn{2}{c}{CRF}&& \multicolumn{2}{c}{Softmax}&& \multicolumn{2}{c}{CRF}&& \multicolumn{2}{c}{Softmax}\\ \cline{2-3}\cline{5-6}\cline{8-9}\cline{11-12}
 & F1 & Acc. && F1 & Acc. && F1 & Acc. && F1 & Acc.
\\\hline 
SE+SemiCRF~\cite{DBLP:conf/ijcai/LiuCGQL16} & 97.3 & -- && -- & -- && {\bf 96.8} & -- & -- & --
\\\hline
WCC embeddings & \multirow{2}{*}{97.8} & \multirow{2}{*}{--} && \multirow{2}{*}{--} & \multirow{2}{*}{--} & & \multirow{2}{*}{96.0} & \multirow{2}{*}{--} && \multirow{2}{*}{--} & \multirow{2}{*}{--}\\ 
+ CRF~\cite{DBLP:conf/emnlp/ZhouYZHDC17} & & && & && & &&  \\\hline
WE + CNN & \multirow{2}{*}{98.0} & \multirow{2}{*}{--} && \multirow{2}{*}{--} & \multirow{2}{*}{--} & & \multirow{2}{*}{96.5} & \multirow{2}{*}{--} && \multirow{2}{*}{--} & \multirow{2}{*}{--}\\ 
+ CRF~\cite{DBLP:conf/ijcnlp/WangX17} & & && & && & && \\\hline
CE + BiLSTM  & \multirow{2}{*}{--} & \multirow{2}{*}{--} &&\multirow{2}{*}{98.1}& \multirow{2}{*}{--} && \multirow{2}{*}{--} & \multirow{2}{*}{--} &&\multirow{2}{*}{96.1} & \multirow{2}{*}{--} \\ ~~+ Softmax~\cite{DBLP:conf/emnlp/MaGW18}& & && & && & &&  \\\hline
BERT & {\bf 98.4} & {\bf 99.0} &&{\bf 98.4} & {\bf 99.0}&& {96.5} & {\bf 97.6} && {\bf 96.5} & {\bf 97.6} 
\\\bottomrule
\end{tabular}
\end{table}
\noindent{\bf Data.}  We evaluate BERT on two benchmark datasets, PKU and MSR, from the second International Chinese Word Segmentation Bake-off~\cite{DBLP:conf/acl-sighan/Emerson05}.  The statistics of the datasets are shown in Table~\ref{tb:dataset}. 

\noindent{\bf Evaluation.}  The standard word F1 measure~\cite{DBLP:conf/acl-sighan/Emerson05} are used to evaluate segmentation performances.  We additionally compute accuracy to evaluate the performance in more aspects.

\noindent{\bf Setting and Setup.}  The experiments are run on a server with 40 cores of Intel Xeon CPU E5-2630 v4 @ 2.20GHz and 128G memory under Linux and the models are trained on one GPU with 12G memory of NVIDIA TITAN Xp graphical card, which totally consists of four GPUs.  

${\rm BERT}_{\rm BASE}$ trained with the Chinese corpus is adopted as the inital model, which consists of 12 BERT layers, the hidden size being 768, the number of self-attention heads being 12, and totally around 110M parameters.  In fine-tuning the model, we adopt ADAM~\cite{DBLP:journals/corr/KingmaB14} as the optimizer.  The learning rate is set to 2e-5.  The maximum sequence length is set to 128.


\subsection{Main Results}
Table~\ref{tb:main_rs} lists the state-of-the-art results from recently applied neural network based models, together with the performance of ${\rm BERT}_{\rm BASE}$ funetuning.  It is observed that CRF and Softmax attain the same performance, where in the MSR dataset, both classifiers achieves the best performance in all models while in the PKU dataset, Softmax attains the best performance and CRF achieves competitive performance among the compared models.  In terms of the Softmax classifier, the finetuning ${\rm BERT}_{\rm BASE}$ can further improve +0.3 F1 score and +0.4 F1 score on the MSR and the PKU dataset, respectively.  

\begin{figure*}[!hbt]
\centering
\subfigure[MSR]{\includegraphics[width=0.45\textwidth]{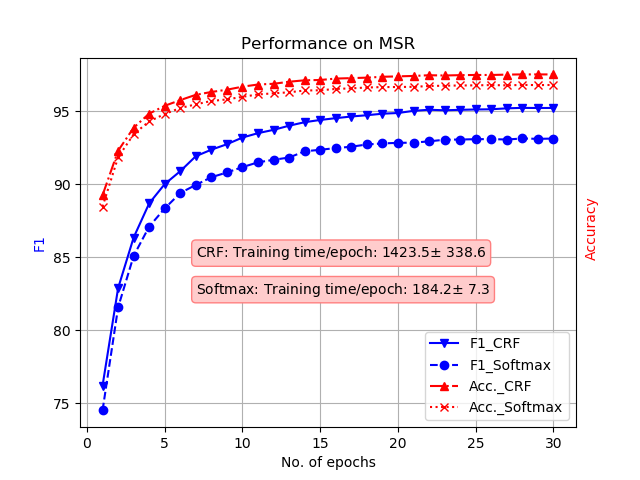}\label{fig:MSR}}
\subfigure[PKU]{\includegraphics[width=0.45\textwidth]{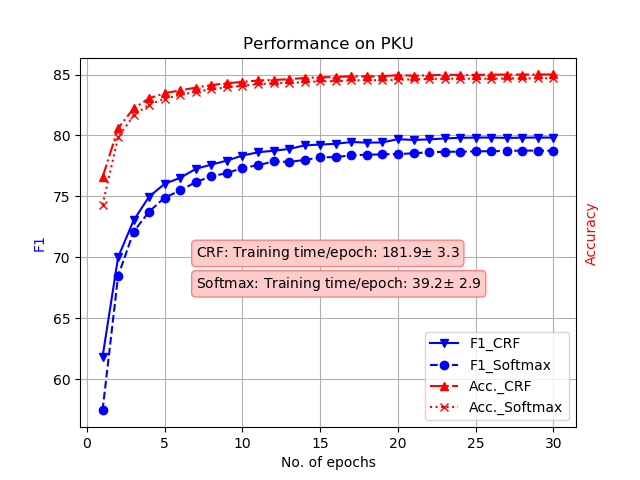}\label{fig:PKU}}
\caption{Ablation study of BERT fine-tuning on only the first/embedding layer.\label{fig:classifiers_comp}}
\end{figure*}

\subsection{Effect of Classifiers}
We investigate the effect of classifiers in both datasets.  In Fig.~\ref{fig:classifiers_comp}, we report the results on fine-tuning the features only from the first layer, i.e., the embedding layer.  The results show that 
\begin{compactitem}
\item The performance of both classifiers increases gradually and converges as the number of epochs increases.  Overall, CRF attains better performance than Softmax when only fine-tuning on one layer. 
\item By observing the results in Table~\ref{tb:layers}, we can find that the gap between CRF and Softmax becomes smaller and they attain the same when the size of layers is 12.   These results imply that when the extracted features are not sufficient, a more complicate classifier (CRF) may help the performance.  
\item We also notice that the time cost per epoch for CRF is much larger than that for Softmax.  It is about 7.7 times (1423.5 seconds vs. 184.2 seconds) in the MSR dataset and 4.6 times (181.9 seconds vs. 39.2 seconds) in the PKU dataset, respectively.  When the BERT model size becomes larger, we can extract more sufficient features.  The results show that the gap of the performance between CRF and Softmax becomes insignificant.  Hence, we would recommend Softmax as the final classifier due to its simplicity. 
\end{compactitem}

\begin{table}
\begin{center}
\caption{Ablation study on BERT with different number of layers (denoted by \#L) on MSR and PKU datasets. \label{tb:layers}}
\begin{tabular}{@{~}r@{~}|@{~}c@{~~}c@{~}c@{~}c@{~~}c@{~}c|@{~}c@{~~}c@{~}cc@{~~}c@{~}}
\toprule
\multirow{3}{*}{\#L}  & \multicolumn{5}{c}{MSR} && \multicolumn{5}{c}{PKU} \\\cline{2-12}
& \multicolumn{2}{c}{CRF}&& \multicolumn{2}{c}{Softmax}&& \multicolumn{2}{c}{CRF}&& \multicolumn{2}{c}{Softmax}\\ \cline{2-3}\cline{5-6}\cline{8-9}\cline{11-12}
 & F1 & Acc. && F1 & Acc. && F1 & Acc. && F1 & Acc.
\\\hline 
1 & 95.2 & 97.5 &&  93.1 & 96.8 && 79.8 & 85.0 && 78.8 & 84.7\\\hline
3 & 97.5 & 98.6 && 97.3 & 98.6 && 84.6 & 88.2 && 83.8 & 87.3\\\hline
6 & 98.2 & 98.9 && 98.1 &98.9 && 85.6 & 88.5 && 86.9 & 89.9 \\\hline
12  & {98.4} & {99.0} &&{98.4} & {99.0} && {96.5} & {97.6} && {96.5} & {97.6} 
\\\bottomrule
\end{tabular}
\end{center}
\end{table}

\subsection{Effect of Model Size}
We explore the effect of model size on fune-tuning the BERT model with different number of layers.  Due to the out-of-memory issue, we set the batch size to 384, 128, 64, and 32 when the number of layers is 1, 3, 6, and 12, respectively.  

From the results in Table~\ref{tb:layers}, we observe that 
\begin{compactitem}
\item The performance increases gradually when the model size (the number of layers) increases.  
\item CRF usually attains better performance than Softmax when the number of layers is small, except for the case when $L=6$ in the PKU dataset.  We conjecture that it is because the PKU dataset is a relative small dataset with a larger out-of-vocabulary (OOV) rate, which makes the model not well trained. 
\end{compactitem}

\begin{table}\caption{Ablation study on BERT with a feature-based approach on MSR and PKU datasets.  The activations from the specified layers are combined and fed into a two-layer BiLSTM, without updating the weights in BERT. \label{tb:BERT_BiLSTM}}
\begin{center}
\begin{tabular}{@{~}l@{~}|@{~}c@{~~}c@{~~}c@{~~}c@{~~}c@{~~}c@{~~}|@{~~}c@{~~}c@{~~}c@{~~}c@{~~}@{~~}c@{~~}}
\toprule
\multirow{3}{*}{Layers}  & \multicolumn{5}{c}{MSR} && \multicolumn{5}{c}{PKU} \\\cline{2-12}
& \multicolumn{2}{c}{CRF}&& \multicolumn{2}{c}{Softmax}&& \multicolumn{2}{c}{CRF}&& \multicolumn{2}{c}{Softmax}\\ \cline{2-3}\cline{5-6}\cline{8-9}\cline{11-12}
 & F1 & Acc. && F1 & Acc. && F1 & Acc. && F1 & Acc.
\\\hline 
Finetune All & {98.4} & {99.0} &&{98.4} & {99.0} && {96.5} & {97.6} && {96.5} & {97.6} \\\hline
First Layer (Embedding) & 95.0 & 97.1 && 95.1 & 97.1 && 91.9 & 94.7 && 91.1 & 94.2 \\
Second-to-Last Hidden & 96.6 & 97.9 && 95.5 & 97.3 && 94.6 & 96.1 && 94.3 & 95.8 \\
Last Hidden & 96.4 & 97.7 && 95.0 & 97.1 && 82.4 & 85.4 && 93.8 & 95.5 \\
Sum Last Four Hidden & 96.8 & 98.0 && 96.6 & 98.0 && 95.0 & 96.4 && 94.9 & 96.4 \\
Concate Last Four Hidden & 96.9 & 98.0 && 96.7 & 98.0 && 95.1 & 96.5 && 94.9 & 96.4\\
Sum All 12 Layers & 97.1 & 98.2 && 96.9 & 98.2 && 95.2 & 96.6 && 95.1 & 96.6
\\\bottomrule
\end{tabular}
\end{center}
\end{table}

\subsection{Feature-based Results}
We also evaluate how well BERT performs in the feature-based approach by generating ELMo-like~\cite{DBLP:conf/naacl/PetersNIGCLZ18} pre-trained contextual representation on the CWS task.  To do this, we apply the activations from one or more layers {\em without} fune-tuning any parameters of BERT.  These contextual embeddings are used as input to a randomly initialized two-layer BiLSTM before the classification layer.  From the results reported in Table~\ref{tb:BERT_BiLSTM}, we observe that
\begin{compactitem}
\item The best performance is attained by summing representations from all 12 hidden layers of the pre-trained Transformer and it is 1.3 and 1.5 behind the F1 attained by fune-tuning all 12 hidden layers in the MSR dataset while the gap is 1.3 and 1.4 in the PKU dataset.  The results also demonstrate the advantage of BERT for feature-based approaches.
\item The performance using the second-to-last hidden layer is usually better than that using the last hidden layer.  This implies the activations in the last hidden layers do not fit for the final downstream task.
\item The worst performance is attained when CRF applies on the activations of the last hidden layer in the PKU dataset.  This again demonstrates that CRF is not suitable when the training set is small. 
\end{compactitem}

\subsection{Ablation Study}
In order to understand the results obtained by BERT, we also randomly select some errors from the MSR and the PKU test set and manually analyze them.

\begin{CJK}{UTF8}{gbsn}
Similar to the observation in~\cite{DBLP:conf/emnlp/MaGW18}, in the MSR test set, BERT treats 抽象概念 (abstract concept) as 抽象 (abstract) 概念 (concept), respectively, because 抽象 (abstract) has appeared 30 times as a word in the MSR training set.  Different from the observation in~\cite{DBLP:conf/emnlp/MaGW18}, in terms of the word related 权 (right/power), in the MSR test set, BERT only makes a mistake for the case of 统治权 (reign power) and separates it as 统治 (reign) and 权 (power), respectively.  For other cases, ``审批 (vetting)  权 (right)", ``建筑 (construction) 权 (right)", ``领导 (leader) 权 (power)", BERT correctly segments the words as the labeled data, which shows the consistence of BERT in segmenting the word, ``统治权 (reign power)".  In the PKU test set, BERT will group ``关税权 (tariff right)" ``贸易权 (trade right)" ``航行权 (navigation right)", ``诉权 (just claim)", rather than the corresponding separating words, ``关税 (tariff)" and ``权 (right)", ``贸易 (trade)" and ``权 (right)", ``航行 (navigation)" and ``权 (right)", ``诉 (just)" and ``权 (claim)", in the labeled test set, respectively.  The results show that BERT consistently segments the words as the same criterion in the training set, rather than the inconsistence in the test set.  In the PKU test set, BERT segments the words, ``有权有势 (having power and having influence)" and ``位高权重 (paramount and powerful)", which is much better than the manually labeled words, ``有权 (having power)", ``有 (having)", ``势 (influence)"; and ``位 (position)", ``高 (high)", ``权 (power)", ``重 (weighty)".   

In terms of the word related to ``县 (county)" in~\cite{DBLP:conf/emnlp/MaGW18}, in the PKU test set, BERT makes no significant difference in segmenting the related words.  Only two cases are to divide one word into two words, e.g., ``堆龙德庆县 (Dui Long De Qing County)" is divided by `堆龙 (Dui Long)" and ``德庆县 (De Qing County)", and ``县区 (county district)" is divided by ``县 (county)" and ``区 (district)".  Meanwhile, three cases are to combine two words into one word, e.g., ``市县 (city and county)" and ``级 (level)" are combined into ``市县级 (the level of city and county)"; ``县 (county)" and ``政府 (government)" are combined into ``县政府 (county government)"; ``先进 (advanced)" and ``县 (county)" are combined into ``先进县 (advanced county)".  We feel that the combination makes the words more compact.

In the MSR test set, ``县 (county)" will be separated from the words, e.g., ``穷 (poor) 县 (county)", ``县 (county) 人行 (the People's Bank of China)", ``县 (county) 消委会 (the consumer council)".  In the BERT prediciton, they correspond to ``穷县 (poor county)", ``县人行 (the People's Bank of China in county)", and ``县消委会 (the consumer council in county)"， respectively.  BERT only makes a significant mistake  in segmenting this sentence, ``本报发表了记者在山东茌平县采写的调查报告 (The newspaper published a survey report written by reporters in Chiping County, Shandong Province.)".  
\begin{compactitem}
\item The sentence is manually segmented as ``本报 (the newspaper) / 发表 (publish) / 了 (ed, past tense) / 记者 (reporters) / 在 (in) / 山东 (Shandong Province) / 茌平县 (Chiping County) / 采写 (written) / 的 (of) / 调查 (survey) / 报告 (report)".  
\item However, the BERT result is ``本报(the newspaper) / 发表 (publish) / 了 (ed, past tense) / 记者 (reporters) / 在 (in) / 山东 (Shandong Province) / [UNK]平县 ([UNK]ping County) / 采写 (written) / 的 (of) / 调查 (survey) / 报告 (report)".
\item Though this example is not affected, the unknown token, [UNK], usually makes BERT misunderstand the whole sentence and needs additional postprocessing.  
\end{compactitem}
\end{CJK}

\begin{CJK}{UTF8}{gbsn}
\begin{table}\caption{Idoms in the MSR test set are predicted separately by BERT. \label{tb:MSR_idoms_BERT_sep}}
\begin{center}
\begin{tabular}{p{0.11\textwidth}p{0.35\textwidth}||@{~}p{0.53\textwidth}} 
\toprule
{Gold} & & BERT 
\\\hline
神经衰弱 & neurasthenia & 神经 nerve / 衰弱 weak \\\hline
若有所思 & thoughtful & 若 if / 有所 have sth. / 思 think \\\hline
重男轻女 & patriarchal & 重 treasure / 男 boys / 轻 dispise / 女 girls \\\hline
崖崖畔畔 & cliffside & 崖 cliff / 崖 cliff / 畔  side / 畔 side \\\hline
男女平等 & gender equality & 男女 men and women / 平等 equality \\\hline
人定胜天 & people will win the day  & 人 people / 定 ensure / 胜 win 天 / day  \\\hline
另眼相看 & regard with special attention & 另 another / 眼 eye / 相看 stare at each other  \\\hline
不可偏废 & not negligible & 不可 cannot / 偏 partial / 废 abandon \\\hline
一好百好 & A good for all  & 一好百 a good hundred / 好 good \\\hline
外引内联 & introduce investment from abroad and establish lateral ties at home & 外 outer / 引 introduce / 内 inside / 联 unite  \\\hline
以丰补歉 & storing when harvest and making up when deficiency  & 以 take / 丰 rich / 补 make up / 歉 poor
\\\bottomrule
\end{tabular}
\end{center}
\end{table}
\end{CJK}

\begin{CJK}{UTF8}{gbsn}
By exploring other differences in the segmentation results, we observe that they lie in segmenting the idoms.  We list them in Table~\ref{tb:MSR_idoms_BERT_sep}-\ref{tb:PKU_idoms_BERT_whole}, respectively, and make the following observations and conjecture:
\begin{compactitem}
\item All the words in Table~\ref{tb:MSR_idoms_BERT_whole} do not appear in MSR training set.  We conjecture that they come from the out-domain resource trained in BERT. 
\item In Table~\ref{tb:PKU_idoms_BERT_whole}, the words, ``银装素裹", ``不懈努力", ``假冒伪劣", ``至关重要", ``难以为继", ``受益匪浅", ``蔚为壮观", and ``证据确凿", are labeled inconsistently in the training set and the test set of the PKU dataset.  Obviously, BERT fits to the training set and makes different prediction in the test set.  For other words, ``徘徊不前”, ``天真无邪", ``倾囊相助", ``喜中有忧", and ``心知肚明", do not appear in the training set and may come from the out-domain resource trained in BERT. 
\end{compactitem}  
\end{CJK}  

\begin{CJK}{UTF8}{gbsn}
\begin{table}[H]
\caption{Idoms in the MSR test set predicted as a whole by BERT but labeled separately. \label{tb:MSR_idoms_BERT_whole}}
\begin{center}
\begin{tabular}{p{0.11\textwidth}p{0.35\textwidth}||@{~}p{0.53\textwidth}} 
\toprule
BERT & & {Gold}   
\\\hline
新春佳节 & Spring Festival  & 新春 new spring / 佳节 festival  \\\hline
胜券在握 & victory in the grip & 胜券 confidence in the victory / 在握 holding \\\hline
整装待命 & ready to stand by & 整装 get one's things ready / 待命 await orders \\\hline
东升西落 & rise in the east and  set in the west  & 东 east / 升 rise / 西 west / 落 set \\\hline
圆缺盈亏 & wax and wan & 圆 circle / 缺 lack / 盈亏 wax and wan \\\hline
乱采滥伐 & deforestation & 乱 disordered / 采 mine / 滥伐 deforestation\\\hline
稳中求进 & seek improvement in stability  & 稳 steady / 中 middle / 求 seek / 进 improvment \\\hline
心有余力 & have spare energy  & 心 heart / 有余 have a surplus / 力 force
\\\bottomrule
\end{tabular}
\end{center}
\end{table}
\end{CJK}

\begin{CJK}{UTF8}{gbsn}
\begin{table}
\caption{Idoms in the PKU test set are predicted separately by BERT.  \label{tb:PKU_idoms_BERT_sep}}
\begin{center}
\begin{tabular}{p{0.11\textwidth}p{0.32\textwidth}||@{~}p{0.55\textwidth}} 
\toprule
{Gold} & & {BERT} 
\\\hline
落落大方 & natural and graceful & 落落 fall / 大方 generous \\\hline
长夜漫漫 & long night & 长夜 long night / 漫漫 very long \\\hline
取信于民 & win the trust of the people & 取信于 win the trust / 民 people \\\hline
杀人灭口 & murder sb. to prevent divulgence of one's secrets & 杀人  murder / 灭口 do away with a witness \\\hline
刑讯逼供 & use torture to coerce a statement & 刑讯 inquisition by torture / 逼供 extort a confession  \\\hline
匡扶正义 & uphold justice & 匡扶 uphold / 正义 justice 
\\\bottomrule
\end{tabular}
\end{center}
\end{table}
\end{CJK}

\section{Conclusion}
\label{sec:conclusion}
In this paper, we conduct extensive experiments to investigate the effect of BERT in solving the CWS task.  Several oberservations are found from the results:
\begin{compactitem}
\item BERT can slightly improve the performance of the CWS task.  More specifically, in terms of the F1 score achieved by Softmax, there is +0.3 and +0.4 gain for the MSR dataset and the PKU dataset, respectively. 
\item When applying sufficiently learned features, CRF and Softmax attain the same performane.  However, Softmax is more favorate due to low time cost.  
\item The performance of BERT increases gradually as the model size increases.  
\item The features extracted by BERT can be also good candidates for other neural network models. 
\item The analyzed results on the idoms in the PKU dataset help us finding the labeled inconsistence issue in the dataset.  For such prediction errors, it is impossible to correct. 
\end{compactitem}
There are several promising research directions related to our work.  
\begin{compactitem}
\item First, our current implementation is not robust to handle multilingual sentences, which consist of both Chinese characters and English words.  It is practical to design new mechanisms to handle them.  
\item Second, OOV is a critical issue because it will yield an unknown token, which confuses BERT to segment the Chinese words.  It seems that training a new BERT model with more Chinese resource is a potential solution.  
\item Third, the current work aims at solving in-domain CWS.  It is promising to explore effective ways of adapting the trained models to new domains, e.g., social media, which consists of short text and special tokens. 
\end{compactitem}

\begin{CJK}{UTF8}{gbsn}
\begin{table}
\caption{Idoms in the PKU test set predicted as a whole by BERT but labeled separately. \label{tb:PKU_idoms_BERT_whole}}
\begin{center}
\begin{tabular}{p{0.11\textwidth}p{0.33\textwidth}||@{~}p{0.56\textwidth}} 
\toprule
BERT & & {Gold}   
\\\hline
银装素裹 & clad in silvery white  & 银装 silver / 素 plain / 裹 wrap  \\\hline
不懈努力 & unrimitting efforts  & 不懈 unrimitting / 努力 efforts \\\hline
假冒伪劣 & forged and fake commodity & 假冒 counterfeit / 伪劣 fake \\\hline
至关重要 & of great importance & 至关 pretty / 重要 importance 
\\\hline
难以为继 & unsustainable & 难以 difficult / 为继 for succession\\\hline
受益匪浅 & benefit a lot & 受益 benifit / 匪 not / 浅 shallow \\\hline
徘徊不前 & not squatting  & 徘徊 hover / 不 not / 前 preceding \\\hline
天真无邪 & innocent and pure  & 天真 innocent / 无 not have / 邪 evil \\\hline
蔚为壮观 & spectacular & 蔚为 to afford / 壮观 spectacularity \\\hline
倾囊相助 & give one's all to help sb. & 倾 pour out / 囊 bag / 相助 help \\\hline 
喜中有忧 & no joy without annoy & 喜 happy / 中 middle / 有 have / 忧 worry \\\hline 
一动一静 & one move and one quiet & 一 one / 动 move / 一 one / 静 quiet \\\hline 
闻风而止 & smell the wind & 闻 smell / 风 wind / 而 moreover / 止 stop\\\hline 
心知肚明 & sth. that I already know & 心 heart / 知 know / 肚 stomach / 明 understand \\\hline 
证据确凿 & irrefutable evidence & 证据 evidence / 确凿 irrefutable
\\\bottomrule
\end{tabular}
\end{center}
\end{table}
\end{CJK}

\section*{Acknowledgment}
The work described in this paper was partially supported by the Research Grants Council of the Hong Kong Special Administrative Region, China (Project No.~UGC/IDS14/16).

\bibliographystyle{splncs04}
\bibliography{ref/nlp,ref/cv,ref/tech}

\end{document}